\documentclass[letterpaper]{article} 
\usepackage{aaai2026}  
\usepackage{times}  
\usepackage{helvet}  
\usepackage{courier}  
\usepackage[hyphens]{url}  
\usepackage{graphicx} 
\usepackage{natbib}  
\usepackage{caption} 
\usepackage{amsfonts} 
\usepackage{algorithm}
\usepackage{algorithmic}

\usepackage{newfloat}
\usepackage{listings}
\DeclareCaptionStyle{ruled}{labelfont=normalfont,labelsep=colon,strut=off} 
\floatstyle{ruled}
\newfloat{listing}{tb}{lst}{}
\floatname{listing}{Listing}
\usepackage{amsmath}
\title{RLSLM: A Hybrid Reinforcement Learning Framework Aligning Rule-Based Social Locomotion Model with Human Social Norms}
\author{
    Yitian Kou\textsuperscript{\rm 1}\equalcontrib, Yihe Gu\textsuperscript{\rm 2}\equalcontrib, Chen Zhou\textsuperscript{\rm 2}\textsuperscript{\rm 3}\equalcontrib, Dandan Zhu\textsuperscript{\rm 1}\thanks{Corresponding Author.}, Shuguang Kuai\textsuperscript{\rm 2}\textsuperscript{\rm 4}\textsuperscript{\rm 5}\footnotemark[2]
}
\affiliations{
    \textsuperscript{\rm 1}School of Computer Science and Technology, East China Normal University, Shanghai, China\\
    \textsuperscript{\rm 2}Shanghai Key Laboratory of Mental Health and Psychological Crisis Intervention, Key Laboratory of Brain Functional Genomics (Ministry of Education and Shanghai), Institute of Brain and Education Innovation, School of Psychology and Cognitive Science, East China Normal University, Shanghai, China.\\
    \textsuperscript{\rm 3}School of Psychology and Neuroscience, University of Glasgow, Glasgow, UK\\
    \textsuperscript{\rm 4}NYU-ECNU Institute of Brain and Cognitive Science, Shanghai, China.\\
    \textsuperscript{\rm 5}Shanghai Center for Brain Science and Brain-Inspired Technology, Shanghai, China.\\
    \{yitian.kou, yhgu\}@stu.ecnu.edu.cn, chen.zhou@glasgow.ac.uk, ddzhu@mail.ecnu.edu.cn, sgkuai@psy.ecnu.edu.cn

}

\usepackage{bibentry}

\begin{document}

\maketitle

\begin{abstract}
Navigating human-populated environments without causing discomfort is a critical capability for socially-aware agents. While rule-based approaches offer interpretability through predefined psychological principles, they often lack generalizability and flexibility. Conversely, data-driven methods can learn complex behaviors from large-scale datasets, but are typically inefficient, opaque, and difficult to align with human intuitions. To bridge this gap, we propose \textbf{RLSLM}, a hybrid \textbf{R}einforcement \textbf{L}earning framework that integrates a rule-based \textbf{S}ocial \textbf{L}ocomotion \textbf{M}odel, grounded in empirical behavioral experiments, into the reward function of a reinforcement learning framework. The social locomotion model generates an orientation-sensitive social comfort field that quantifies human comfort across space, enabling socially aligned navigation policies with minimal training. RLSLM then jointly optimizes mechanical energy and social comfort, allowing agents to avoid intrusions into personal or group space. A human-agent interaction experiment using an immersive VR-based setup demonstrates that RLSLM outperforms state-of-the-art rule-based models in user experience. Ablation and sensitivity analyses further show the model’s significantly improved interpretability over conventional data-driven methods. This work presents a scalable, human-centered methodology that effectively integrates cognitive science and machine learning for real-world social navigation.
\end{abstract}

\begin{figure*}[!ht]
    \centering
    \includegraphics[width=0.95\textwidth]{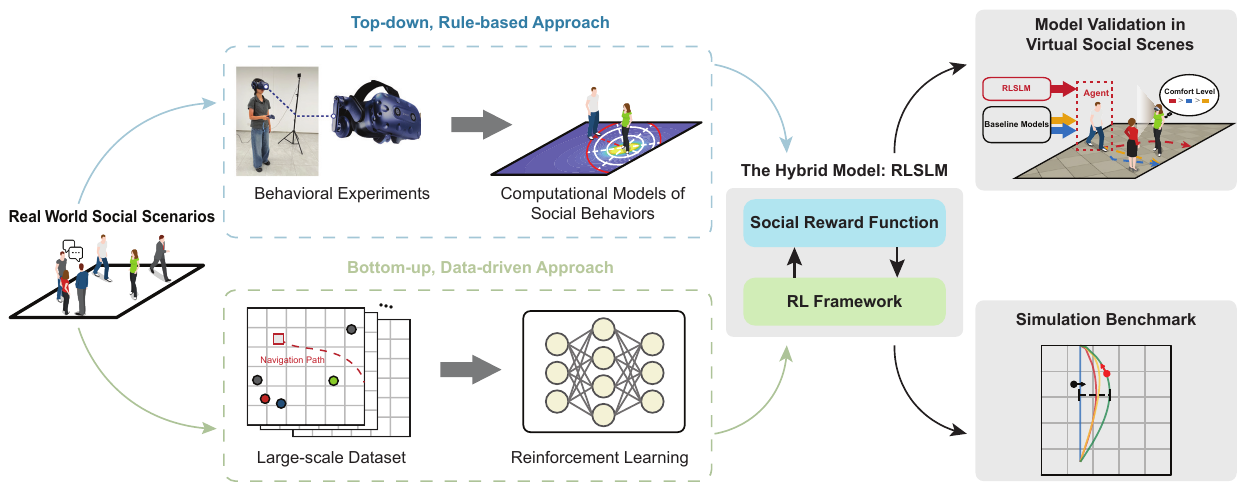}
    \vspace{-4pt}
    \caption{Methodology overview. The hybrid model of RLSLM combines the top-down, rule-based approach which develops computational models of human social behaviors from well-controlled lab experiments and bottom-up data-driven approach which formulates the reinforcement learning framework based on large-scale dataset of real-world social scenarios. The hybrid model first encodes human behavioral patterns into a social reward function, which is then used to train the policy within a reinforcement learning framework. The trained model is subsequently validated through human-agent interaction studies and simulations.}
    \label{fig:figure1}
\end{figure*}

\section{Introduction}

Moving around human-populated environments without causing discomfort is an essential requirement for social agents, since they are widely engaged in human-agent interaction \cite{sheridan2016human}. Such socially-aware navigation entails consideration of multiple social factors and remains a highly challenging problem \cite{francisPrinciplesGuidelinesEvaluating2025}.

Existing work on socially-aware navigation can be broadly classified into two categories, rule-based and data-driven. Rule-based approaches typically adopt models with identified variables and interpretable, quantifiable principles, like proxemics \cite{doi:10.1177/1729881418776183} and velocity \cite{kim2015brvo}, either derived from social psychology or manually designed. Although these models show strength in interpretability and low computational overhead, they are often (1) difficult to quantify precisely, (2) limited in generalizability across environments, and (3) less flexible, which may lead to unnatural behaviors like oscillatory paths \cite{kretzschmarSociallyCompliantMobile2016}, ultimately constraining their real-world applicability.

Meanwhile, data-driven methods, such as reinforcement learning (RL) \cite{wangMultiRobotCooperativeSociallyAware2024} and imitation learning \cite{karnanSociallyCompliAntNavigation2022}, have enabled agents to emulate human navigation behaviors based on large-scale human trajectory datasets \cite{kapoorSocNavGymReinforcementLearning2023,terryPettingZooGymMultiAgent2021} or simulation environments \cite{mansoSocNav1DatasetBenchmark2020,tsoiSEANSocialEnvironment2020,vuongHabiCrowdHighPerformance2024}. Although these approaches have achieved promising results, they are (1) highly dependent on the quality of the dataset, (2) expensive to train, and (3) often lack interpretability or alignment with human intuitions. With insufficient prior knowledge to guide the training, data-driven methods are often inefficient and prone to pitfalls.

Therefore, an important question arises: can these two approaches be integrated to develop models that are efficient, adaptable, and interpretable—while remaining aligned with real-world human social behavior? To address this, we propose RLSLM, a hybrid framework that integrates a computational social locomotion model derived from psychological research \cite{ZhouMCHCYK22} into the reward structure of an RL agent. Based on well-controlled behavioral experiments, the rule-based social locomotion model computes an orientation-sensitive, asymmetric discomfort field that covers the entire navigation area, with higher field values indicating a greater amount of discomfort that the agent may cause to others when passing that point. By incorporating this rule-based model into a multi-objective RL framework to jointly minimize mechanical energy and social discomfort, we enable the agent to learn complex socially aligned rules within a small number of training epochs, such as avoiding invasion of personal space and social groups.

We further compared RLSLM with two rule-based models using human comfort ratings. The results demonstrate that our framework significantly outperforms these baselines in terms of users’ comfort.

In summary, this work contributes:
\begin{itemize}
    \item \textbf{A novel hybrid RL framework} that integrates a psychologically grounded social locomotion model into reinforcement learning, combining the interpretability and prior knowledge of rule-based methods with the adaptability and expressiveness of data-driven approaches. This framework is potentially generalizable and applicable in other scenarios with similarly scarce data. 
 \item \textbf{Performance breakthrough in user comfort}: RLSLM achieves a mean comfort rating of 4.21/5, significantly outperforming the best rule-based baseline ($\Delta \text{rating} = 1.12$, Bonferroni corrected post-hoc comparisons, $P<0.001$). This establishes a new Pareto frontier in the trade-off between comfort and efficiency.
\end{itemize}

\section{Related Work}
\subsection{Incorporating Social Rules in Navigation}
Recent studies have explored the incorporation of social rules into navigation algorithms. The design of social rules modules is mostly driven by intuition, dataset statistics, or physical modeling of human path planning. Static properties like the proper radius of personal space are often determined by intuition and experience in previous studies \cite{gongCognitionPrecognitionFutureAware2025}. Qualitative navigation decisions (e.g. passing on the left or right when encountering others) and trajectory features that facilitate path prediction can be learned from real-world pedestrians datasets \cite{kretzschmarSociallyCompliantMobile2016}. To support dynamic path planning in human-populated scenarios and avoid collision, physical-based models like the social force model have been developed to simulate particle-like motion of the crowd \cite{helbingSocialForceModel1995,shiomiSociallyAcceptableCollision2014}, often corresponding to intuitive geometric relations instead of real pedestrians’ movements \cite{chenSociallyAwareMotion2018}.
In conclusion, although a great number of navigation studies have taken social rules into account, most of them are not quantitatively grounded in human behavioral experiments, which can lead to the generation of unnatural paths \cite{chenDecentralizedNoncommunicatingMultiagent2016}. Notably, a recent study determines social-aware parameters (e.g. neighbor distance) through a user experiment, in which participants are presented with simulated navigation videos and asked to report their perceived social comfort \cite{beraSociallyInvisibleRobot2018a}. However, this approach fails to account for the subtle characteristics of path planning, and third-person user studies may yield unrealistic feedback due to issues of ecological validity. To address the aforementioned limitations, we propose a hybrid framework that directly embeds findings from cutting-edge psychological research into training workflows, and test our model performance in an immersive VR experiment with high ecological validity.

\subsection{Learning-Based Approaches to Social Navigation}

Deep learning has significantly advanced trajectory forecasting by enabling data-driven modeling of complex social interactions. GAN-based methods such as SocialGAN~\cite{Gupta2018} and SoPhie~\cite{Sadeghian2019} capture multimodal behaviors by generating diverse plausible futures. Graph-based models like Social-STGCNN~\cite{Mohamed2020} leverage GNNs to model spatial interactions among pedestrians. Transformer-based models like  STAR~\cite{yu2020spatio} and STPOTR~\cite{10160538} have further improved performance through long-range temporal modeling and attention-based interaction. These models implicitly learn conventions from data, but offer limited interpretability and little control over compliance with social rules or physical rules. RL provides a more natural and interpretable framework for integrating such rules into trajectory generation. 
To our knowledge, no prior work leverages reinforcement learning to explicitly encode and optimize human-centered social navigation  constraints. We bridge this gap by casting navigation as reward-driven optimization with interpretable social influence modeling.

\begin{figure*}[t]
\centering
\includegraphics[width=0.95\textwidth]{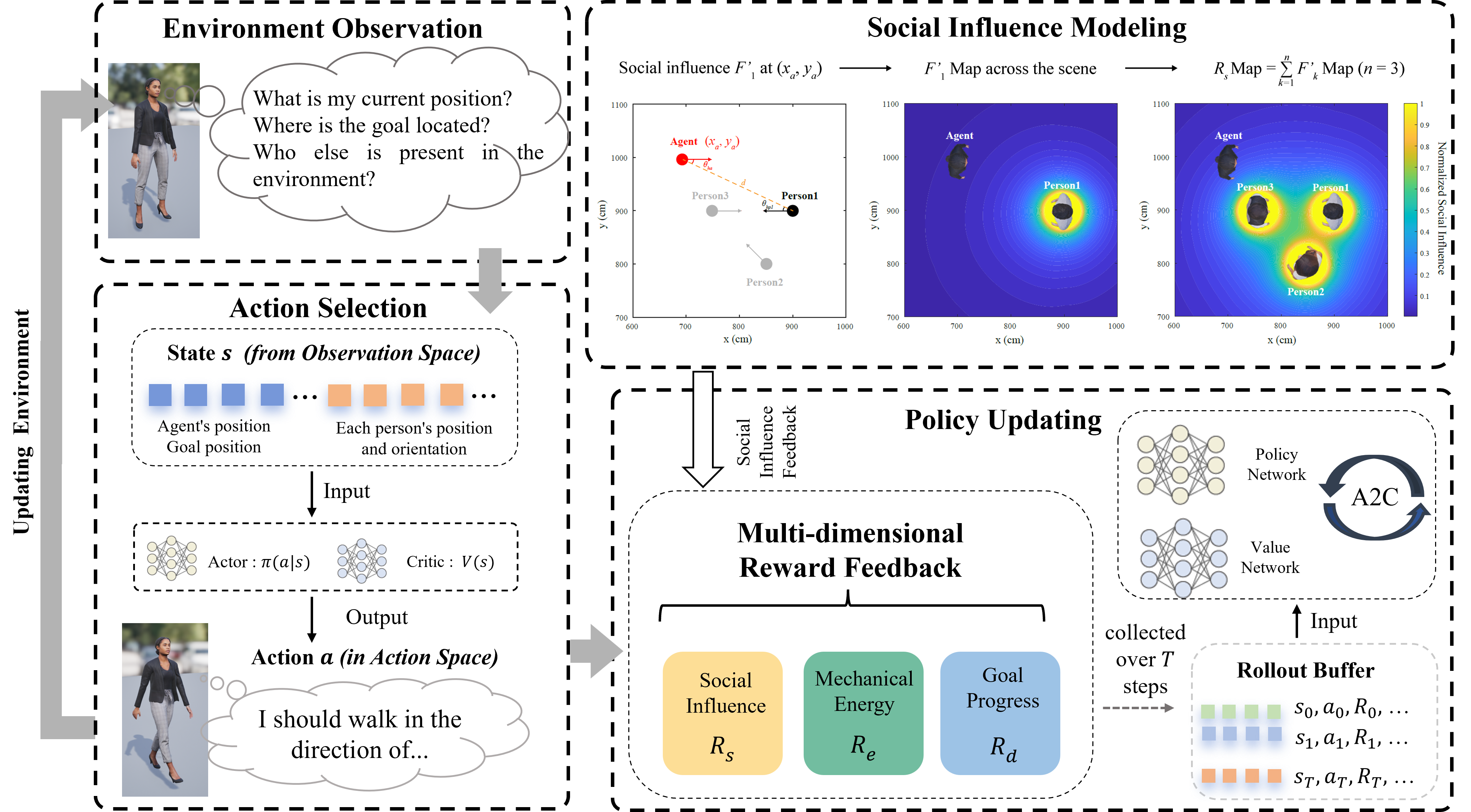} 

\caption{Overview of RLSLM framework. RLSLM integrates social influence modeling with reinforcement learning to guide an agent’s movement in environments shared with humans. The framework follows a three-stage decision-making loop (gray arrow), and once the environment is updated based on the agent's action, the cycle begins again with a new observation.}

\label{fig.Algorithm Framework}
\end{figure*}

\section{Method}

In this section, we introduce a reinforcement learning framework for socially-aware navigation, which integrates environment observation to capture agent states and social cues, an actor-critic network for effective action selection, and a multi-dimensional feedback mechanism.

As illustrated in Figure~\ref{fig.Algorithm Framework}, the agent's decision-making pipeline is composed of three core components: environment observation, action selection, and policy updating. In the following subsections, we describe each module in detail and explain how the feedback signals are formulated to promote socially compliant and efficient navigation behavior.

\subsection{Agent Decision-Making Process}
The agent’s behavior is guided by a three-stage decision-making process: perceiving the environment, selecting actions based on the learned policy, and continuously updating the policy through reinforcement learning.

\textbf{Environment Observation }
The environment observation module captures the agent’s perception of the surroundings. Similar to how humans perceive social environments, the agent’s observation space includes not only its own position, but also the relative positions and orientations of surrounding individuals. These features are concatenated into a structured input vector $s_t$. For a scenario involving \( n \) individuals, the observation vector at timestep \( t \) is represented as \( s_t \in \mathbb{R}^{3n + 2} \).

\textbf{Action Selection }
To learn a policy that guides the agent’s decision-making, we adopt a deep reinforcement learning framework based on an actor-critic architecture. Specifically, the policy is represented by a stochastic function $\pi(a_t | s_t)$, which denotes the probability of selecting action $a_t$ given the current state $s_t$.
The actor network models the policy $\pi(a_t | s_t)$ and generates a distribution over possible actions, supporting a trade-off between exploration and exploitation. Concurrently, the critic network estimates the value function $V(s_t)$, which predicts the expected return from state $s_t$. This actor-critic architecture allows the agent to optimize its policy through trial-and-error interaction with the environment, gradually improving its navigation performance in socially complex settings.

\textbf{Policy Updating }
At each timestep, the agent observes the current state and selects an action accordingly. The output action $a_t$ represents a navigation command indicating a movement direction. The agent then executes this action, causing the environment to transition to a new state $s_{t+1}$, and receives a reward signal for learning. 
We utilize the Advantage Actor Critic algorithm (A2C) \cite{mnih2016asynchronousmethodsdeepreinforcement} to jointly train the actor and critic. The actor is updated to maximize the expected return by increasing the probability of actions with high advantage estimates, while the critic is trained to minimize the temporal difference (TD) error between successive value predictions. Through continuous interactions with the environment and iterative policy updates, the agent learns to generate socially compliant and energy-efficient behaviors that achieve the task objective.

\subsection{Multi-dimensional Feedback Mechanism}
During social navigation, approaching other individuals in the scene tends to cause greater discomfort (i.e., experiencing social influence from others), while taking a detour from them consumes more mechanical energy. To balance social influence and mechanical energy consumption and emulate human-like path planning behavior, we design a multi-dimensional feedback mechanism that provides rewards and penalties based on three key factors: mechanical energy expenditure, progress toward the goal and social influence.

\textbf{Mechanical Energy }
We assume that the agent moves a fixed distance $l$ at each timestep, and does not consider the mechanical energy required for turning, which implies that the mechanical energy expenditure per step remains constant. To penalize excessive energy usage and encourage efficient motion, we introduce a negative reward component \( R_e\) defined as:
\begin{equation}
R_e(s_t) = -\alpha,
\end{equation}
where $\alpha$ is a constant representing the estimated mechanical energy consumed per step. This term ensures that the agent is incentivized to reach the goal using the minimal number of steps, thereby promoting energy-efficient navigation.

\textbf{Goal Progress }
To encourage the agent to make progress toward the goal, we introduce a positive reward component \( R_d\), which is proportional to the reduction in distance between consecutive timesteps:
\begin{equation}
R_d(s_t,s_{t-1}) = \frac{D_{t-1} - D_t}{l},
\end{equation}
where \( D_{t-1} \) and \(D_t\) represent the distances from the agent to the destination at the previous and current timesteps, respectively.

\textbf{Social Influence }
To encourage socially-aware behavior in the agent, we incorporate insights from prior psychological research~\cite{ZhouMCHCYK22} to quantify social influence. Based on results from behavioral experiments, the social influence of each individual on surrounding space is modeled as an orientation-sensitive, asymmetric field, with higher field values indicating greater discomfort. The field model comprises three components: a heading-relevant social component (HRSC), a heading-irrelevant social component (HISC), and a collision avoidance component (CAC). Given the relative locations and orientations of the agent and surrounding persons, the add-up influence (with persons index $1\sim k$) $R_s$ imposed on the agent at $s_t$ can be computed as follows:

\begin{equation}
R_s(s_t) = \sum\limits_{k}F'_k,
\end{equation}

\begin{equation}
F' = \min\left(\frac{F}{K},\ 1\right),
\end{equation}

\begin{equation}
F = \frac{I_{\text{agent}} \times I_{\text{person}}}{d^2},
\end{equation}

\[
I_{\text{human}} = m \times f(\theta_h) + n + c \times I_{\text{CA}}, 
\]
\begin{equation}
\quad \text{where} \quad 
f(\theta_h) = 
\begin{cases}
\cos(\theta_h), & \cos(\theta_h) \geq 0 \\
0, & \text{otherwise}
\end{cases}, \quad \quad 
\end{equation}

\begin{equation}
I_{\text{CA}} = \frac{ab}{\sqrt{a^2 \cos^2(\theta) + b^2 \sin^2(\theta)}},
\end{equation}
where $F$ represents the original social influence field value and $d$ represents the distance between the agent and $\text{person}_k$ at $s_t$. The individual social influence of the agent and $\text{person}_k$, $I_\text{agent}$ and $I_\text{person}$, is calculated using $I_{\text{human}}$ with different fitted parameters $m$ and $n$ (for the agent, $m_a= 0.321$, $n_a = 0.856$; for the $\text{person}_k$, $m_p = 0.438$, $n_p = 0.630$). Specifically, $m$ represents the contribution of HRSC and $n$ represents HISC. $\theta_h$ represents the angle between the facing direction and the line connecting agent--$\text{person}_k$. $I_{\text{CA}}$ represents CAC, with $a$ and $b$ ($a = 0.285$, $b = 0.175$) estimated by measuring the average cross-section of human body (approximated as an ellipse), and $\theta$ represents the angle between the line connecting agent--$\text{person}_k$ and the long arm of the ellipse. The free parameter $c$ adjusted the relative ratio ($c = 1.430$) of $I_{\text{CA}}$. To prevent extreme $F$ values, original social influence $F$ is standardized with an upper limit $K$ ($K = 10.180$) fitted by behavioral data. The values of all these parameters are set according to the prior work by Zhou et al.

Given that the agent is deemed to have reached the destination if its distance to the endpoint is less than a predefined threshold. At the terminal timestep \( t = T \), the agent receives a reward \( r_T = +C \) if it successfully reaches the destination, or \( r_T = -C \) otherwise (exceeding step limit or moving out of bounds). Overall, the return $G$ is calculated as the sum of discounted rewards:
\begin{equation}
    G = \sum_{t=0}^T \gamma r_t ,
\end{equation}
where $\gamma$ is the discount factor, and $r_t$ is the reward at timestep $t$, which is composed of three components and a terminal reward, defined as follows:


\begin{equation}
r_t = 
\begin{cases} 
R_d(s_t, s_{t-1}) + R_e(s_t) + \sigma R_s(s_t), \quad {\text{for } 0 < t < T} \\
\pm C, \quad \text{for } t = T
\end{cases}
\end{equation}
where $\sigma$ denotes the weight of social influence. In our experiments, the parameters are set as follows: \( \gamma = 0.9 \), \( \sigma = 0.5 \), \( C = 500 \), and \( \alpha = 1 \). Additional training details are available in the supplementary material.

\section{Dataset Creation}
To support reproducible research and facilitate benchmarking in socially-aware navigation, we establish a VR-based human-agent interaction dataset. 
This dataset is designed as a benchmark environment for evaluating social comfort under controlled yet immersive conditions. Our dataset comprises a diverse set of simulated VR scenarios featuring varied human placements and orientations. The environment is implemented using Unreal Engine 5.4 and supports a variable number of virtual humans with configurable positions and orientations, covering a range of common social patterns such as face-to-face blocking, group passage, and asymmetric crowd formations. The dataset also includes user annotations on social comfort collected through immersive VR experiments. 
\begin{figure}[!t]
    \centering
    \includegraphics[width=0.9\linewidth]{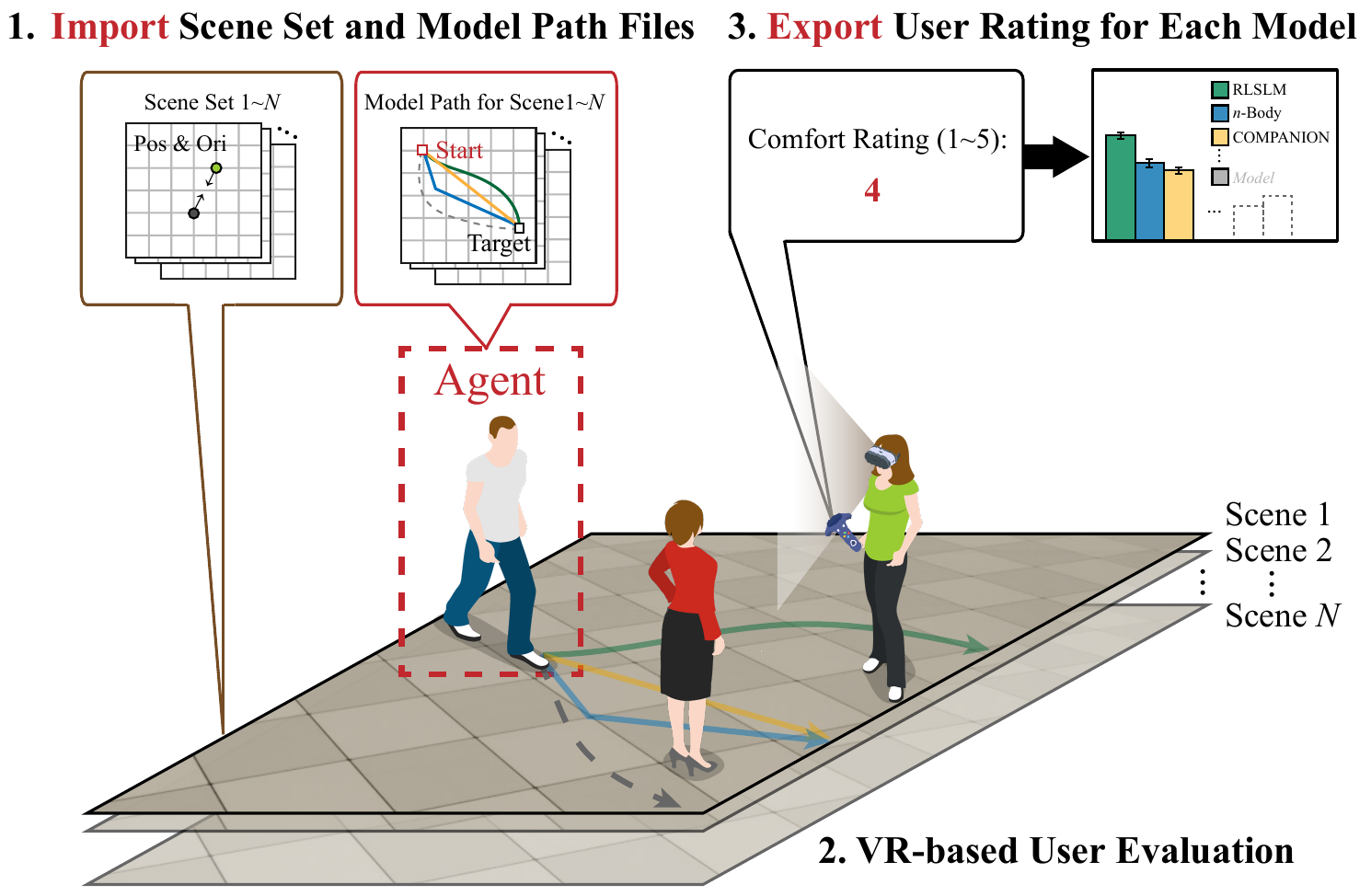} 
    \caption{Overview of our VR-based user evaluation pipeline. (1) For each scenario, we import a set of human layouts (positions and orientations) and corresponding navigation trajectories generated by different models. (2) The participant views these simulated interactions in an immersive first-person VR environment, observing the agent’s movement among virtual humans. (3) After each trial, the participant provides a comfort rating (1–5), which is recorded and aggregated across models for quantitative comparison.
}
    \label{fig:vr-eval}
\end{figure}

The process of constructing our immersive evaluation dataset is illustrated in Figure~\ref{fig:vr-eval}. For each scene, we define static human layouts and import precomputed model trajectories. Participants then experience these scenarios in VR and rate the agent's navigation behavior. Implementation details and code are provided in the supplementary materials.
To facilitate reproducibility and further research, we provide public access to both the dataset and the VR evaluation pipeline. Beyond evaluation, this dataset can also serve as a reusable benchmark for future studies on human-centered navigation evaluation.

\section{Experiments}

\begin{figure*}[!ht]
\centering
\includegraphics[width=0.95\textwidth]{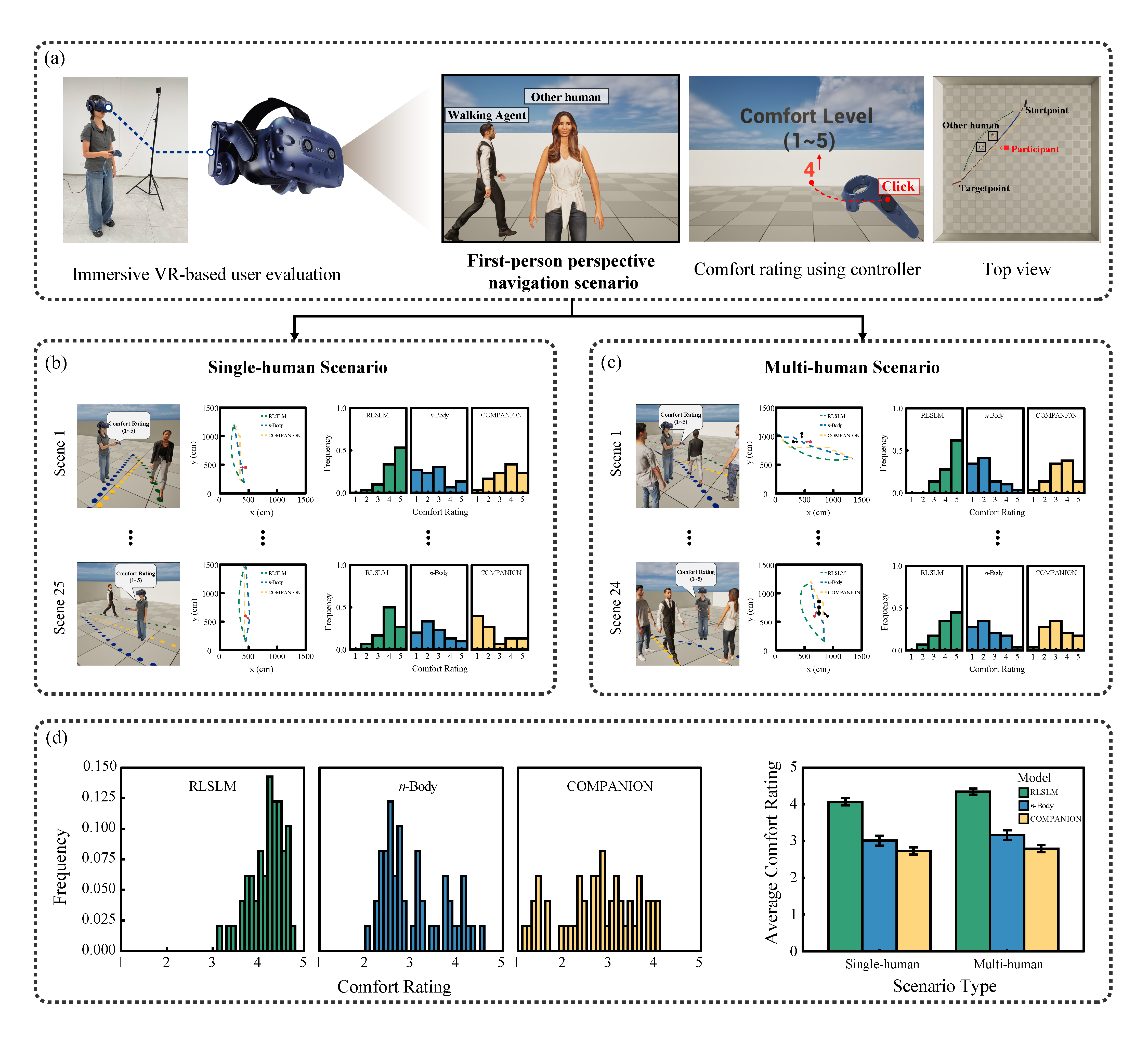} 
\vspace{-20pt}
\caption{Comfort Rating Analysis via VR-Based User Study. (a) illustrates the VR experiment setup, where participants rate their comfort level (1–5) in both single- and multi-human interaction scenarios.
(b) and (c) shows the trajectories of each model (RLSLM, $n$-Body, and COMPANION) from a top-down view. We selected two representative cases from both scenarios for presentation; the complete results are provided in the supplementary material.
(d) presents the comfort rating distributions for each model under both scenarios, comparing the average comfort ratings of three models across both single- and multi-human interaction scenarios.}
\label{fig.VR}
\end{figure*}

\subsection{Experimental Setup}
To assess the generalizability of the model, we perform evaluations in both single- and multi-human scenarios, with the latter involving three individuals. In single-human scenarios, we position the human near the straight line connecting the start and goal points to observe the agent’s avoidance behavior when encountering one human. In multi-human scenarios, we arrange two or three individuals in a social interaction state, such as facing each other, to examine whether the agent respects social formations or intrusively passes between interacting individuals. The experiment is conducted within a confined $15m\times15m$ virtual environment. The agent's step length is fixed at $45 cm$. At each decision step, following an observation of the environment, the agent selects a movement direction and advanced exactly one step in that direction. The episode is considered successful when the Euclidean distance between the agent and the target falls below the length threshold of one step. Due to the discrete stepwise movements of the agent, the resulting trajectories tended to exhibit discontinuities and jaggedness. To enhance spatial continuity, we post-process the trajectories using a Gaussian smoothing filter to improve path smoothness. 

Following prior work \cite{sivashangaran2023autovrlhighfidelityautonomous}, we construct our virtual environment using OpenAI Gymnasium \cite{towers2024gymnasiumstandardinterfacereinforcement} and employ Stable-Baselines3 \cite{stable-baselines3} for policy learning. Comprehensive training configurations and convergence plots are presented in the supplementary material.

\subsection{Human-Agent Interaction Experiment in VR}
To test whether RLSLM better aligns with user experience in human-agent interaction, we conduct a VR-based experiment in which participants are asked to rate their comfort level towards the virtual agents controlled by one of three navigation algorithms: RLSLM, COMPANION \cite{kirbyCOMPANIONConstraintOptimizingMethod2009a}, and $n$-Body \cite{vandenbergReciprocalNBodyCollision2011a}.

\textbf{Participants}
A total of 30 university students and staff (11 males and 19 females aging between 18 and 29) are recruited to participate in this study. All participants have normal or corrected-to-normal vision.

\textbf{Procedure }
We randomly set 50 scenarios in which a virtual agent bypasses one or three static persons along a detour path generated by one of three navigation algorithms, resulting in 150 trials. Notably, one of the multi-human scenarios is found to be repetitive and is thus excluded in the following analysis. This exclusion has no impact on statistical conclusions (a detailed comparison of statistical results is available in the supplementary material). As shown in Figure~\ref{fig.VR} (a), participants experience these scenarios as one of the static persons via an HTC Vive Pro head-mounted display (HTC Corporation; binocular resolution: 2,880×1,600 pixels; refresh rate: 90 Hz; field of view: 110°). In each scenario, participants stand at the designated location and orientation of a randomly selected static person, while the virtual agent walks from the start point to the target point along the pre-generated path. Upon completion of the navigation, participants rate their comfort level (on a scale from 1 to 5, with 5 indicating maximum comfort) using a handheld controller.

\textbf{Result Analysis }
Figure~\ref{fig.VR} (b) and (c) illustrate the trajectories produced by three models. Figure~\ref{fig.VR} (d) presents a detailed analysis of user evaluation data. A significant main effect of model type on comfort level is present (repeated-measures ANOVA, $F_{(2, 58)}=219.589$, $P<0.001$, $\eta_G^2=0.525)$. Rating scores of paths generated by RLSLM are significantly higher than those generated by COMPANION (both in single-human and multi-human scenarios, Bonferroni corrected post-hoc comparisons, $P<0.001$) and $n$-Body (both in single-human and multi-human scenarios, Bonferroni corrected post-hoc comparisons, $P<0.001$). Compared to single-human scenarios, paths generated by RLSLM and $n$-Body receive significantly higher scores in multi-human scenarios (Bonferroni corrected post-hoc comparisons, RLSLM: $P<0.001$, $n$-Body: $P=0.008$), whereas COMPANION does not exhibit this multi-human navigation advantage (Bonferroni corrected post-hoc comparisons, $P=0.251$).

\begin{figure*}[!ht]
    \centering
    \includegraphics[width=0.95\textwidth]{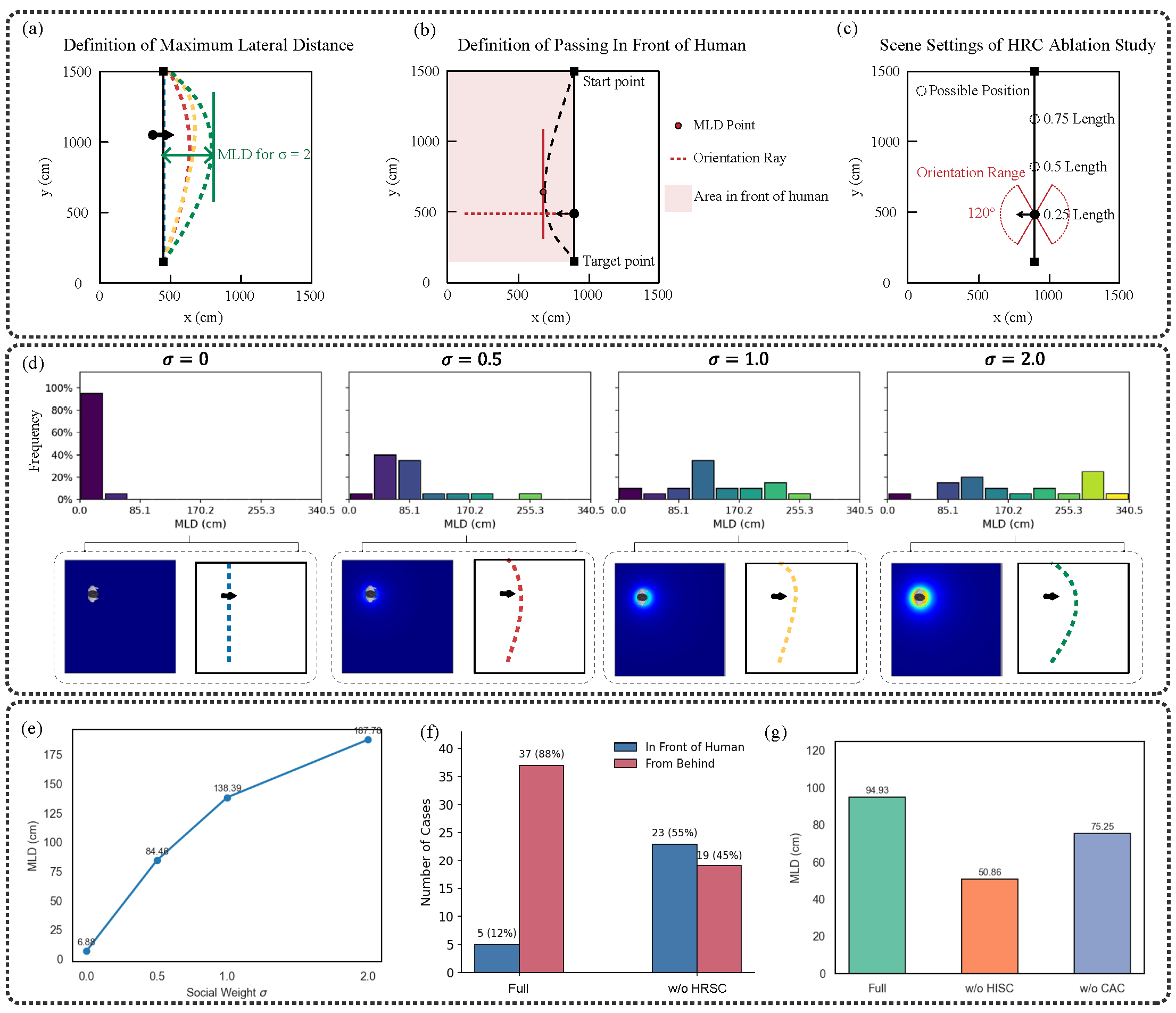}
    \vspace{-4pt}
    \caption{Model validation and ablation analysis.
    \textbf{(a–c)} Definitions and experimental setup.
    \textbf{(d–e)} Effects of varying the social behavior weight $\sigma$: (d) shows trajectory examples and MLD distributions under different $\sigma$ values; (e) reports the corresponding average MLD statistics.
    \textbf{(f–g)} Statistical results from ablation studies of the heading-relevant (f) and heading-irrelevant (g) components of the social influence model. Full experimental details are provided in the appendix.}

    \label{fig:Figure5}
\end{figure*}

\subsection{Interpretable Modeling of Social Behavior Weighting}


To evaluate interpretability and adaptability to social variability, we perform a sensitivity analysis on the social behavior weight $\sigma \in \{0, 0.5, 1.0, 2.0\}$, which modulates the agent's responsiveness to nearby individuals through the social influence field. As shown in Figure~\ref{fig:Figure5} (d, e), higher $\sigma$ values lead to greater lateral deviations, quantified using Maximum Lateral Distance (MLD) defined in Figure~\ref{fig:Figure5} (a). When $\sigma = 0$, the agent strictly follows the shortest path; as $\sigma$ increases, the agent detours more, prioritizing social comfort. At $\sigma = 2.0$, behavior becomes overly conservative. These trends confirm the effectiveness of $\sigma$ in shaping socially aware navigation. Further analysis and results are detailed in the appendix.




\subsection{Ablation Study on Social Influence Components}
We isolate the contributions of the three components in our social influence module: HRSC, HISC, and CAC.

\textbf{Heading-Relevant Component }
In 42 specially-designed single-human scenarios (Figure~\ref{fig:Figure5} (c)), removing HRSC causes the agent to pass in front of humans (defined in Figure~\ref{fig:Figure5} (b)) in 23 cases (57.76\%), compared to only 5 cases using the full model, as shown in (Figure~\ref{fig:Figure5} (f)). This demonstrates that HRSC enables sensitivity to human orientation.

\textbf{Heading-Irrelevant Components }
Measured in 21 single-human scenarios, removing HISC or CAC leads to reduced MLD (Figure~\ref{fig:Figure5} (g)), indicating less stable and less compliant navigation. Qualitative trajectory visualizations and experiment setups are provided in the appendix.

\section{Conclusion}
In this paper, we present RLSLM, a hybrid reinforcement learning framework grounded in empirical behavioral experiments for socially compliant robot navigation in human-shared spaces. By integrating a quantitative, rule-based SLM derived from psychological research into a multi-objective RL formulation, our method enables agents to navigate not only efficiently but also in a manner aligned with human social preferences. Through a combination of mechanical energy minimization, goal-directed progress, and social discomfort reduction, the agent learns socially-aware behaviors that generalize to multi-human scenarios. To evaluate alignment with human perception, we designed an immersive first-person VR evaluation pipeline. Results demonstrate that RLSLM significantly outperforms rule-based baseline models in subjective comfort ratings. Additionally, ablation studies and sensitivity analyses underscore the role of each component in shaping nuanced social behavior and demonstrate RLSLM's improved interpretability over conventional data-driven methods. Our findings highlight a promising interdisciplinary pathway for embedding human social cognition into agent policy learning.

\section{Acknowledgments}
This work is supported by the National Science and Technology
Innovation 2030 Major Program [grant number
2022ZD0205103], the National Natural Science Foundation
of China [grant number T2425028], and the National Natural Science Foundation of China [grant number 62377011].

\bibliography{aaai26}

\appendix
\section*{Appendix}
In this section, we will provide a brief supplementary introduction to the reinforcement learning (RL) procedure, ablation study on social influence components, and full set of trajectories used in our VR-based experiments.

We have open-sourced our example code and UE project for VR Evaluation Pipeline on \url{https://github.com/kouyitian/RLSLM} to facilitate reproducibility.

\subsection*{VR Evaluation Pipeline}

During the evaluation process, each participant experiences the scenario from a fixed first-person perspective within the virtual environment, allowing for consistent spatial perception and controlled evaluation.

At the beginning of each trial, participants are instructed to observe and confirm the positions and orientations of the virtual humans in the environment. After reviewing all nearby humans, they then identify the agent's start and goal positions. After that, a brief auditory cue indicates the beginning of the robot’s movement. As the robot agent begins to move, participants can hear spatialized footstep sounds, which vary in volume based on distance to simulate real-world auditory perception. The robot's trajectory passes near the participant’s virtual avatar, enabling naturalistic evaluation of its social appropriateness.

Once the robot reaches the goal, the trial ends. Participants are then prompted to rate the trajectory based on their subjective experience. Two aspects are assessed: perceived social comfort and perceived path rationality. Ratings are given on 5-point Likert scales as described in Tables~\ref{tab:evaluation-scale}.
\begin{table}[ht]
\centering

\resizebox{\columnwidth}{!}{
\begin{tabular}{c|l}
\hline
\textbf{Score} & \textbf{Evaluation} \\
\hline
1 & Extremely inappropriate: rude, irrational, or unsafe\\
2 & Somewhat inappropriate: impolite or inefficient\\
3 & Neutral: acceptable but unremarkable\\
4 & Generally appropriate: polite and efficient\\
5 & Very appropriate: socially considerate and highly rational\\
\hline
\end{tabular}
}
\caption{Trajectory Evaluation Rating Scale (5-point Likert).}
\label{tab:evaluation-scale}
\end{table}

During the scoring process, the initial score values are randomized; for a given scenario with three methods, the display order of the trajectories corresponding to the three methods is also randomized.

\subsection*{Reinforcement Learning Details}
\begin{figure}[ht]
    \centering
    \includegraphics[width=0.9\linewidth]{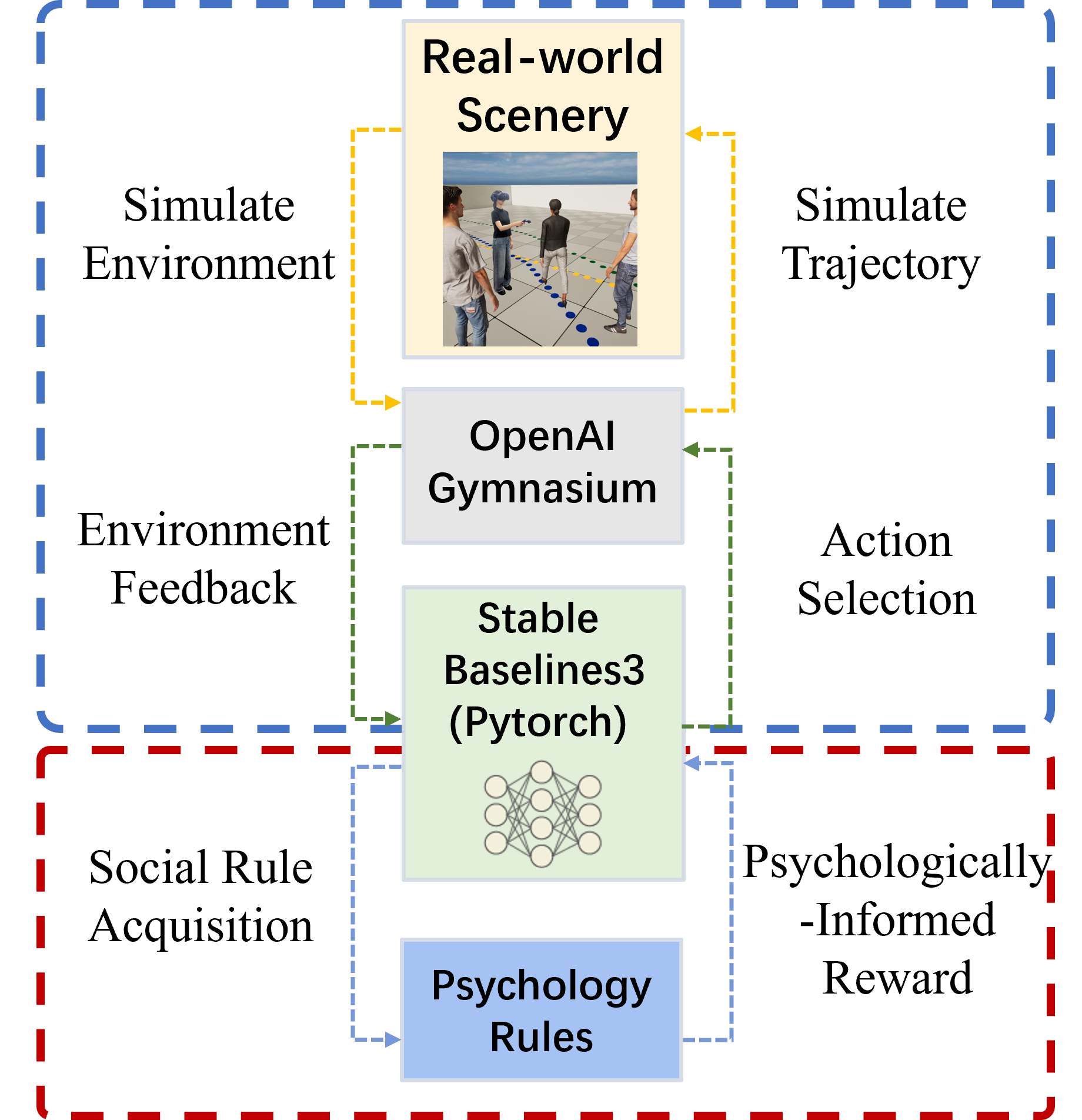} 
    \caption{Schematic of RLSLM architecture.}
    \label{fig:framework}
\end{figure}
Figure~\ref{fig:framework} illustrates the interaction between the environment, learning algorithm, and psychology-driven reward design in our framework. Real-world social scenarios are abstracted into simulated environments via OpenAI Gymnasium. The training process is implemented using Stable Baselines3, where the reward function is guided by psychology-inspired social discomfort rules. The learning agent adapts behavior through training, and psychological insights are encoded into the reward structure. Finally, the learned policy can also serve as a computational tool for psychology research. By modeling how agents internalize and respond to social rules, our framework enables controlled experimentation on the influence of social discomfort, spatial norms, and interaction strategies—offering new insights into human social cognition from a quantitative perspective.

For policy learning, we utilize the Advantage Actor-Critic (A2C) algorithm as implemented in the Stable-Baselines3 library, with RMSprop as the optimizer and a learning rate set to $5 \times 10^{-4}$. Both the actor and critic networks employ a multilayer perceptron (MLP) policy with a symmetric architecture of five layers: 64–128–256–128–64 hidden units. The discount factor $\gamma$ is configured to $0.8$ to balance immediate and long-term rewards. Training runs on an NVIDIA 3090 GPU using CUDA acceleration. All experiments operate under a fixed time budget of 10,000 steps per run.

To monitor learning progress, we log training metrics, tracking episode-level statistics such as average reward and path length. We perform separate training runs for each scenario condition to ensure environment-specific convergence.

\begin{figure}[ht]
\centering
\includegraphics[width=\linewidth]{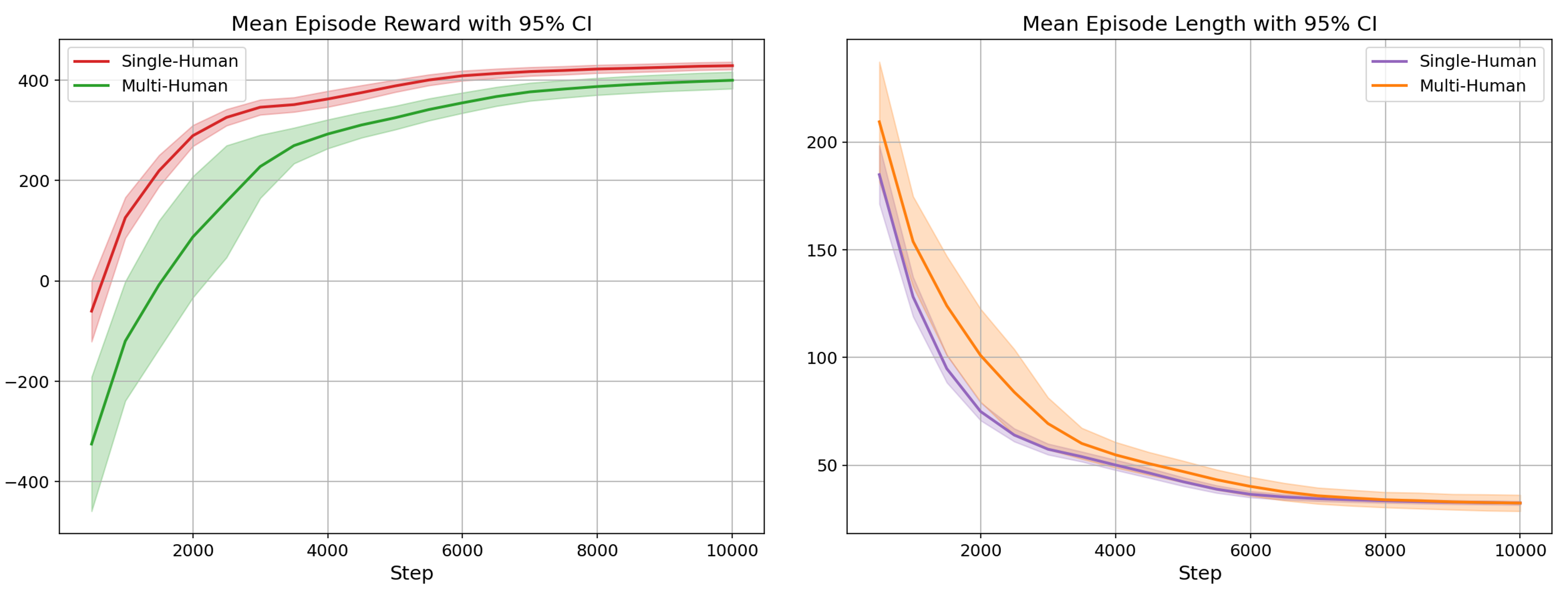}
\caption{
Reinforcement learning performance under single-human and multi-human scenarios. 
\textbf{Left:} Mean episode reward over training steps, with shaded areas indicating 95\% confidence intervals.
\textbf{Right:} Mean episode length across steps, again with 95\% confidence bands. }
\label{fig:rl-training}
\end{figure}

Figure~\ref{fig:rl-training} presents the learning curves under single-human and multi-human setting. In both scenarios, the agent successfully improves its navigation strategy over time. Episode rewards increase steadily, while episode lengths decrease, reflecting more efficient and socially compliant behavior. Although learning in the multi-human scenario converges more slowly due to increased interaction complexity, it still achieves stable performance within 10,000 training steps.

\subsection{Ablation Study on Social Influence Components}
To investigate the effectiveness of social modeling in RLSLM, we perform a systematic ablation study by individually disabling the three social influence components—HRSC, HISC and CAC—within the SLM module, and evaluate their respective contributions to navigation performance.
We design two controlled single-human settings to separately examine the effects of the heading-relevant component (HRSC) and the heading-irrelevant components (HISC and CAC).

\textbf{Heading-Relevant Component }
To observe the relationship between the path chosen by the agent and the orientation of individuals, we design a special experiment involving 42 single-human scenarios. In each scenario, the individual is positioned along the straight line connecting the agent’s start and goal locations, with their orientation forming an angle greater than 30° relative to this line. Among these scenarios, the model without the heading-relevant component (w/o HRSC) selected a path that passed in front of the human in 23 cases (57.76\%), indicating a lack of sensitivity to orientation and a tendency to avoid the individual randomly on either side. In contrast, our full model passed in front of the human in only 5 cases, demonstrating a more consistent awareness of directional social cues. These results underscore the importance of the heading-relevant component in enabling socially aware and direction-sensitive navigation. Implementation details and additional results are provided in the Supplementary Material.

\begin{figure}[h]
    \centering
    \includegraphics[width=0.9\linewidth]{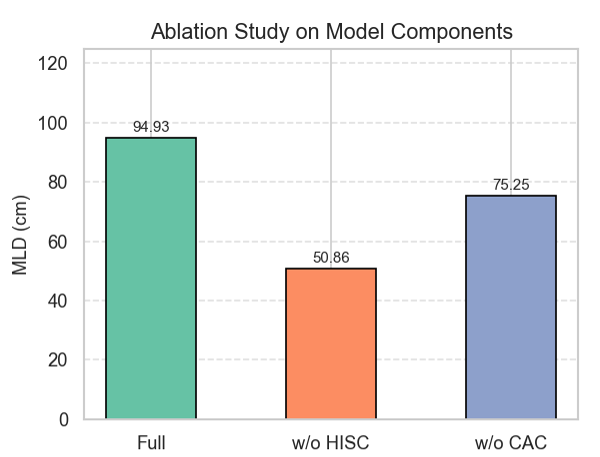} 
    \caption{Ablation study results on the Maximum Lateral Distance (MLD) in the single-human scenario. The comparison includes the full model, a variant without HISC, and a variant without CAC.}
    \label{fig:ablation}
\end{figure}

\textbf{Heading-Irrelevant Components }
We assess the contribution of the heading-irrelevant components by measuring MLD in 21 single-human scenarios. As reported in Figure~\ref{fig:ablation}, removing either component results in a noticeable decline in MLD, suggesting reduced navigation stability. Qualitative trajectory visualizations for both single- and multi-human scenarios are included in the Supplementary Material to further illustrate these effects.

\subsection{Full Trajectory Comparison with Other Methods}
In this section, we present the full set of trajectories used in the VR-based user study to support a comprehensive comparison of path quality across methods.

\begin{figure}[h]
\centering
\includegraphics[width=\linewidth]{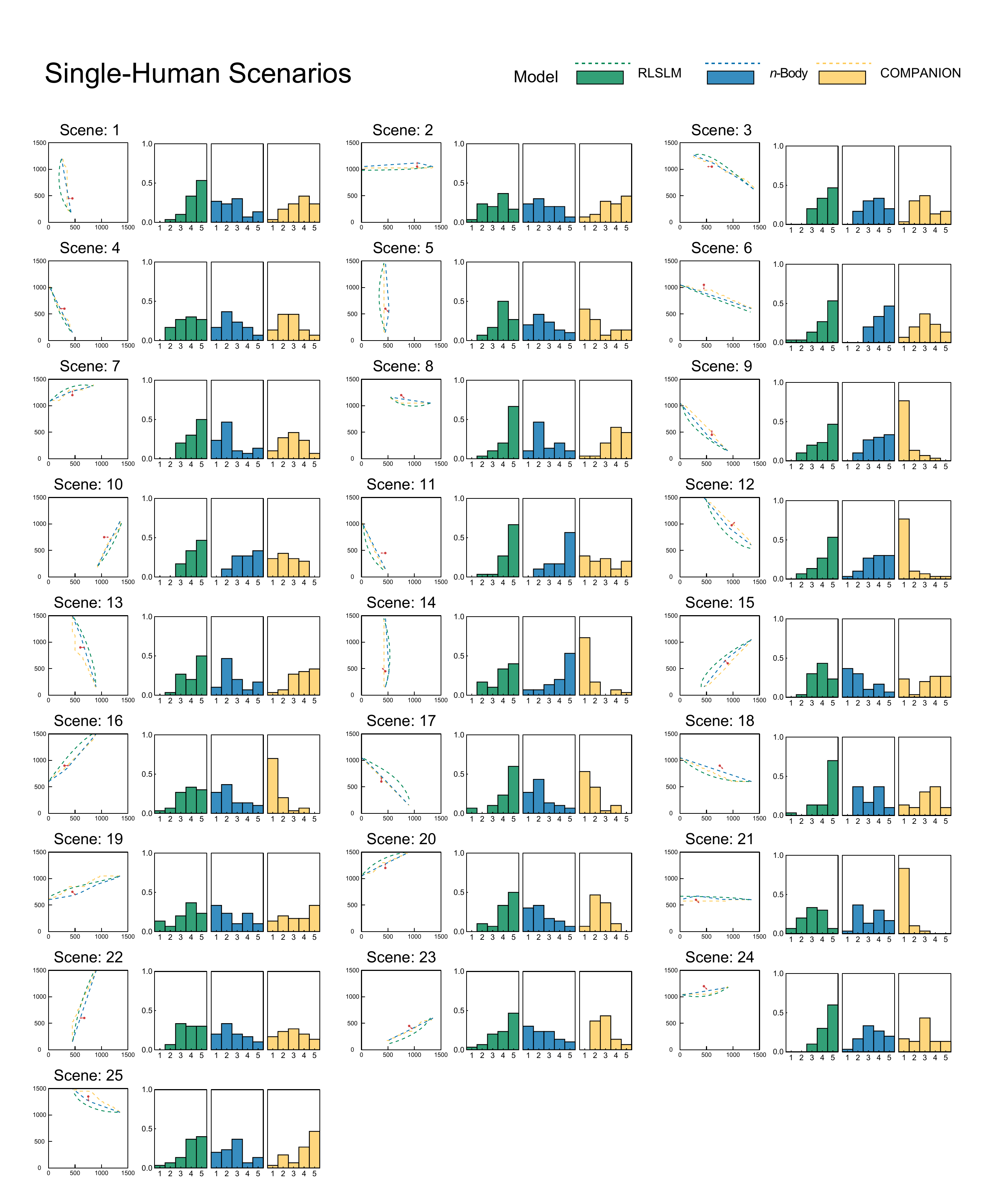}
\caption{Qualitative comparison of navigation performance across 25 single-human scenarios. For each scene, three models are evaluated: \textbf{RLSLM} (green), \textbf{n-Body} (blue) and \textbf{COMPANION} (orange). Each subplot contains (1) a trajectory length distribution (left), and (2) a histogram of user-rated scores (right, Likert scale 1–5) obtained from the immersive VR study. Dashed lines represent predicted trajectory of each model.
}
\label{fig:single-user-eval}
\end{figure}

\begin{figure}[h]
\centering
\includegraphics[width=\linewidth]{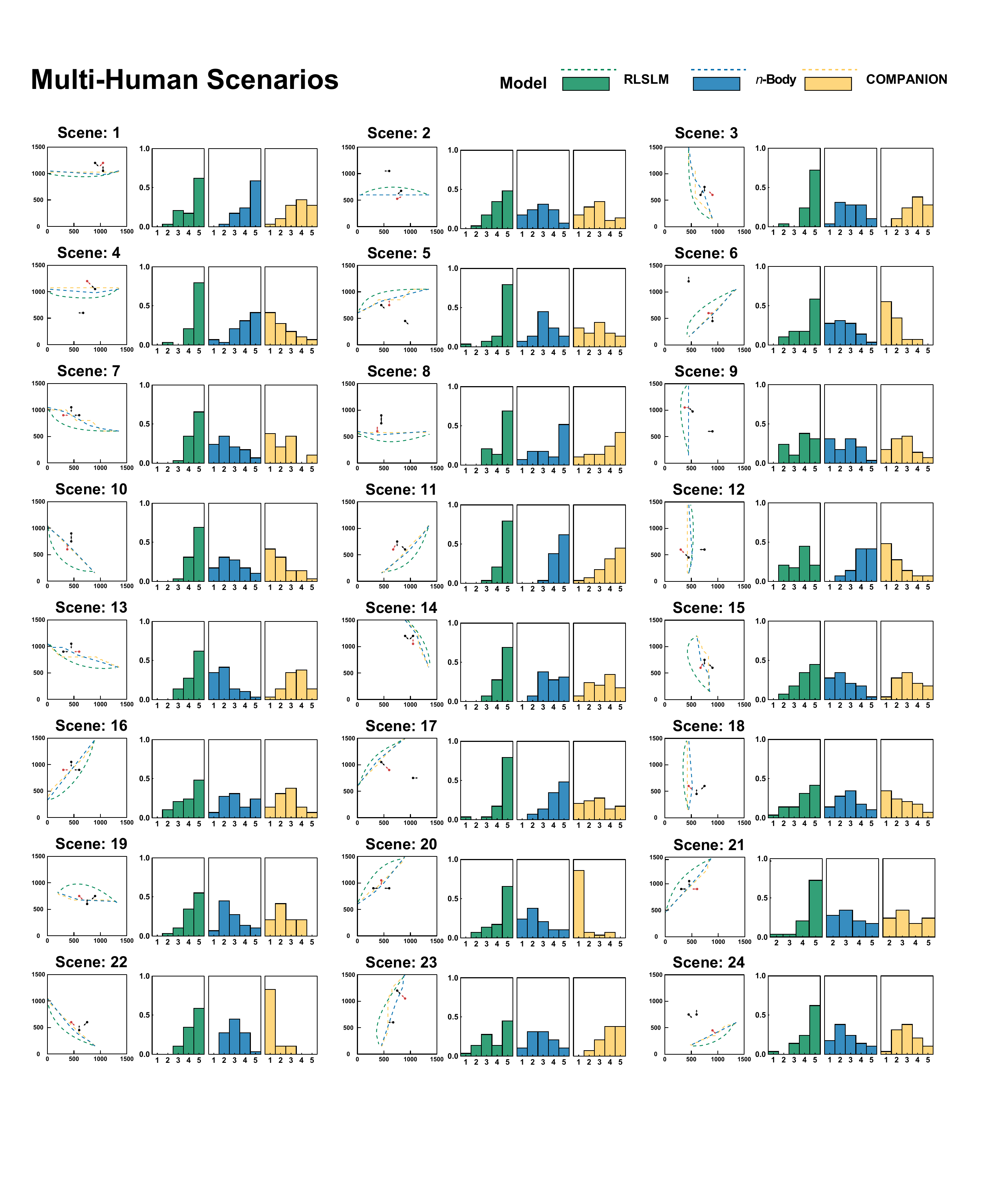}
\caption{
Qualitative comparison of navigation performance across 24 multi-human scenarios. For each scene, three models are evaluated: \textbf{RLSLM} (green), \textbf{n-Body} (blue) and \textbf{COMPANION} (orange). Each subplot contains (1) a trajectory length distribution (left), and (2) a histogram of user-rated scores (right, Likert scale 1–5) obtained from the immersive VR study. Dashed lines represent predicted trajectory of each model.
}
\label{fig:multi-user-eval}
\end{figure}

Figures~\ref{fig:single-user-eval} and~\ref{fig:multi-user-eval} summarize the comparison across 50 scenarios involving a single human and three humans, respectively. All trajectories shown were used in the immersive VR evaluation to assess perceived comfort and social appropriateness. For each scenario, we report both the predicted trajectories and the distribution of user rating scores (on a 5-point Likert scale), evaluated across three navigation models. Dashed lines indicate the predicted trajectory of each model.

 As shown in Figure~\ref{fig:single-user-eval}, our model consistently receives higher comfort ratings while producing more moderate and stable trajectories. In the more complex multi-human settings shown in Figure~\ref{fig:multi-user-eval}, RLSLM maintains its advantage—achieving higher comfort scores while avoiding overly conservative detours—highlighting its robustness in socially dense environments.

\subsection{Comparison of Statistical Results in Human-Robot Interaction Experiment}

Before/After excluding the repetitive three-human scene (before: 25 scenarios for both one-human and three-human settings; after: 25 scenarios for one-human setting and 24 scenarios for three-human setting):
A significant main effect of model type on comfort level is present (repeated-measures ANOVA, before: $F_{(2, 58)}=228.112$, $P<0.001$, $\eta_G^2=0.534)$; after: $F_{(2, 58)}=219.589$, $P<0.001$, $\eta_G^2=0.525)$. Rating scores of paths generated by RLSLM are significantly higher than those generated by COMPANION (both in single-human and multi-human scenarios, Bonferroni corrected post-hoc comparisons, before/after: $P<0.001$) and n-Body (both in single-human and multi-persons scenarios, Bonferroni corrected post-hoc comparisons, before/after: $P<0.001$). Compared to single-human scenarios, paths generated by RLSLM and n-Body receive significantly higher scores in multi-human scenarios (Bonferroni corrected post-hoc comparisons, before: RLSLM: $P<0.001$, $n$-Body: $P=0.032$; after: RLSLM: $P<0.001$, $n$-Body: $P=0.008$), whereas COMPANION does not exhibit this multi-human navigation advantage (Bonferroni corrected post-hoc comparisons, before: $P=0.931$; after: $P=0.251$).

\end{document}